%% file: acl2023.tex
\pdfoutput=1
\documentclass[11pt]{article}

\usepackage[]{ACL2023}

\usepackage{times}
\usepackage{latexsym}

\usepackage[T1]{fontenc}
\usepackage[utf8]{inputenc}

\usepackage{microtype}

\usepackage{inconsolata}
\usepackage{graphicx}
\usepackage{multirow}
\usepackage{arydshln}
\usepackage{tabularray}
\usepackage{rotating}
\usepackage{float}
\usepackage[T5]{fontenc}
\usepackage[utf8]{inputenc}
\usepackage{lscape}
\usepackage{xcolor}
\usepackage[linesnumbered,ruled,vlined]{algorithm2e}
\usepackage{amsfonts}

\title{\textsc{ViHateT5}: Enhancing Hate Speech Detection in Vietnamese\\With a Unified Text-to-Text Transformer Model}

\author{
    Luan Thanh Nguyen$^{1, 2}$ \\
    $^{1}$Faculty of Information Science and Engineering, University of Information Technology, \\Ho Chi Minh City, Vietnam \\
    $^{2}$Vietnam National University, Ho Chi Minh City, Vietnam \\
    \texttt{luannt@uit.edu.vn}
    }

\begin{document}
\maketitle
\begin{abstract}
Recent advancements in hate speech detection (HSD) in Vietnamese have made significant progress, primarily attributed to the emergence of transformer-based pre-trained language models, particularly those built on the BERT architecture. However, the necessity for specialized fine-tuned models has resulted in the complexity and fragmentation of developing a multitasking HSD system. Moreover, most current methodologies focus on fine-tuning general pre-trained models, primarily trained on formal textual datasets like Wikipedia, which may not accurately capture human behavior on online platforms. In this research, we introduce \textsc{ViHateT5}, a T5-based model pre-trained on our proposed large-scale domain-specific dataset named VOZ-HSD. By harnessing the power of a text-to-text architecture, \textsc{ViHateT5} can tackle multiple tasks using a unified model and obtain state-of-the-art performance on all benchmark HSD datasets in Vietnamese. The experiments also underscore the significance of label distribution in pre-training data on model efficacy. We provide our experimental materials for research purposes, including the VOZ-HSD dataset\footnote{https://huggingface.co/datasets/tarudesu/VOZ-HSD}, pre-trained checkpoint\footnote{https://huggingface.co/tarudesu/ViHateT5-base}, the unified HSD-multitask \textsc{ViHateT5} model\footnote{https://huggingface.co/tarudesu/ViHateT5-base-HSD}, and related source code on GitHub\footnote{https://github.com/tarudesu/ViHateT5} publicly.

% We provide our experimental materials for research purposes, including the VOZ-HSD dataset\footnote{Link provided upon acceptance}, pre-trained checkpoint\footnote{Link provided upon acceptance}, the unified HSD-multitask ViHateT5 model\footnote{Link provided upon acceptance}, and related source code on GitHub\footnote{Link provided upon acceptance}.

\textbf{Warning:} This paper contains examples from actual content on social media platforms that could be considered toxic and offensive.
\end{abstract}

\input{sections/1_intro}
\input{sections/2_relatedwork}
\input{sections/3_ViHateT5}
\input{sections/4_experiments}
\input{sections/5_conclusion}
\input{sections/6_limitations}
\input{sections/7_ethical}
\input{sections/8_acknowledgement}

\bibliography{anthology,custom}
\bibliographystyle{acl_natbib}

\clearpage

\input{sections/9_appendix}

\end{document}

%% file: sections/1_intro.tex
\section{Introduction}
\label{sec:introduction}

% What is hate speech?
Hate speech refers to harmful expression targeting individuals or groups based on their inherent characteristics, potentially inciting violence or discrimination \cite{brown2017hate}. Its detrimental impact on mental well-being includes different levels of anxiety, depression, or stress among affected individuals \cite{ghafoori2019global}. Due to the rise of social media on the internet, where people can easily leave their toxic content that may negatively affect anyone who reads it, the consequences that hate speech brings to use become worse and worse. To address these issues, automatic systems have been explored to detect harmful content online and mitigate its dissemination \cite{gitari2015lexicon,macavaney2019hate,aimeur2023fake}. 

% Language problems
Different languages have unique forms of harmful expressions, necessitating specific text-processing methodologies for developing HSD systems. In the context of English, one of the most prevalent languages, there exist several effective strategies for addressing HSD-related tasks, such as employing machine learning models \cite{abro2020automatic} or deep learning models for identifying harmful content \cite{badjatiya2017deep,zimmerman2018improving}. Furthermore, transfer learning approaches have garnered considerable research interest, showcasing the effectiveness in solving hate speech detection tasks \cite{ali2022hate,mozafari2020bert}. In the case of low-resource languages, numerous studies have been conducted, yielding promising results in tackling this issue \cite{bigoulaeva-etal-2021-cross,nkemelu2022tackling,arango-monnar-etal-2022-resources}.

%%% TO FIX THE POSITION OF THE FIGURE %%%
\begin{figure*}[t!] 
    \centering
    \includegraphics[width=\textwidth]{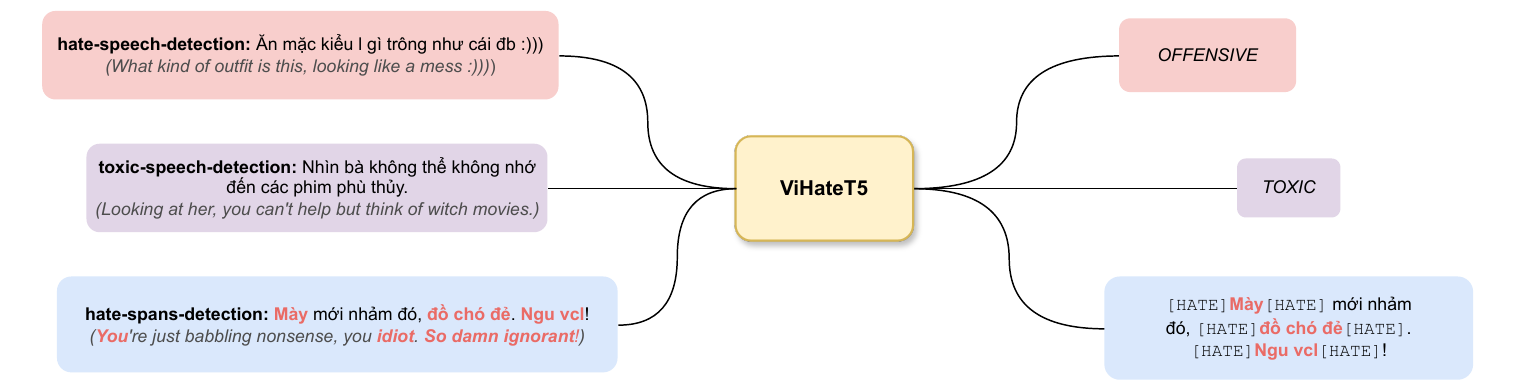}
    \caption{An overview of the unified HSD-multitask \textsc{ViHateT5} model incorporating various prefix tasks tailored for hate speech detection in Vietnamese.}
    \label{fig:ViHateT5-example}
\end{figure*}
%%% TO FIX THE POSITION OF THE FIGURE %%%

% Vietnamese - low-resource processing
Vietnamese, considered a low-resource language, has seen limited research in the field of NLP concerning high-quality datasets and pre-trained models. Recent efforts in hate speech detection tasks based on Vietnamese language characteristics have yielded significant achievements \cite{vu2020hsd,luu2021large,hoang-etal-2023-vihos}. However, current state-of-the-art models only fine-tune general transformer-based models, which may have been pre-trained on formal textual data sources \cite{nguyen-tuan-nguyen-2020-phobert}. Moreover, even pre-trained on social media text, models like ViSoBERT \cite{nguyen-etal-2023-visobert} still necessitate separate fine-tuning for specific tasks, resulting in system fragmentation.

% Paper contributions
This paper presents a novel approach to address the existing limitations of HSD systems in Vietnamese. The contributions of this research are outlined as follows:

\begin{itemize}
    \item A domain-specific model named \textsc{ViHateT5} is presented in this study to address HSD-related problems in the Vietnamese language. Our innovative model was explicitly trained on a specific dataset derived from social media texts called VOZ-HSD with 10M+ comments with generated labels, designed to perform multiple tasks in a unified framework.
    
    \item A unified T5-based model, obtained by fine-tuning the pre-trained \textsc{ViHateT5} model, advances the state-of-the-art performance on all available HSD benchmark datasets in Vietnamese. 

    \item We illustrate our empirical strategy and data preparation to establish a comprehensive model for tackling HSD problems. Moreover, we highlight the significance of data pre-training on our pre-trained model, showing its substantial impact on model performance. 
\end{itemize}

% Paper structure
This paper is structured into distinct sections. Section \ref{sec:related-work} examines relevant prior research on Vietnamese HSD-related tasks. Subsequently, Section \ref{sec:vihatet5} introduces \textsc{ViHateT5}, our text-to-text model, covering its automatically generated pre-training dataset, pre-training methodologies, and fine-tuning for downstream tasks. Section \ref{sec:experiments} presents experimental results obtained by comparing various baseline methods with our proposed \textsc{ViHateT5} model across a range of hate speech detection-related tasks and address discussions. Section \ref{sec:conclusion} concludes the paper with a summary of our findings. Section \ref{sec:limitation} addresses the current limitations of our proposed method, while Section \ref{sec:ethical} provides ethical statements related to our research.

%% file: sections/2_relatedwork.tex
\section{Related Work}
\label{sec:related-work}

% Effectiveness of transfer learning approaches
Since the emergence of the transformer architecture \cite{devlin-etal-2019-bert}, numerous challenges in NLP have been successfully addressed, including tasks related to hate speech detection. Additionally, domain-specific models like HateBERT \cite{caselli-etal-2021-hatebert}, fBERT \cite{sarkar-etal-2021-fbert-neural}, or ToxicBERT\footnote{https://huggingface.co/unitary/toxic-bert} have been introduced. However, there remains a deficiency in hate-speech-focused pre-trained models for low-resource languages like Vietnamese, hindering the effective resolution of HSD tasks.

% Current research about HSD in Vietnamese
Besides, there exist diverse endeavors on HSD tasks, which involve the contribution of large-scale, high-quality datasets, thus facilitating precise research in this domain \cite{luu2021large,nguyen2021constructive,hoang-etal-2023-vihos}. Furthermore, competitions such as the VLSP-2019 shared-task \cite{vu2020hsd}, dubbed Hate Speech Detection on Social Networks, aiming at identifying harmful content on internet-based social media, yielding remarkable outcomes. Additionally, transformer-based models have demonstrated remarkable efficacy across various NLP tasks. The advent of monolingual pre-trained models in Vietnamese, which have been observed to surpass their multilingual counterparts \cite{nguyen-etal-2022-smtce, to2021monolingual}, has paved the way for the creation of precise systems for hate speech detection.

% A unified model of HSD
On the contrary, despite the remarkable performance achieved by current BERT-based approaches, developing separate systems tailored to individual tasks is necessary. Addressing this issue of fragmentation, a unified text-to-text-based model on T5 architecture \cite{raffel2023exploring} such as FT5 for English or mFT5 \cite{ranasinghe2023text} has demonstrated its effectiveness in amalgamating multiple tasks ranging from syllable-level to sentence-level-based HSD challenges. Hence, in this study, we introduce \textsc{ViHateT5}, a pre-trained text-to-text model designed explicitly for the hate speech domain, aiming to streamline complex separate systems while ensuring optimal performance in addressing HSD issues in Vietnamese.

%% file: sections/3_ViHateT5.tex
\section{\textsc{ViHateT5}}
\label{sec:vihatet5}

% Overview
This section reveals the methodologies for the creation of pre-training data, the pre-training techniques utilized, and the fine-tuning procedures undertaken to assemble the unified \textsc{ViHateT5} model.

\subsection{Automated Pre-training Data Creation}
\label{sec:automated-data-creation}

\begin{figure}[H]
    \centering
    \includegraphics[width=0.5\textwidth]{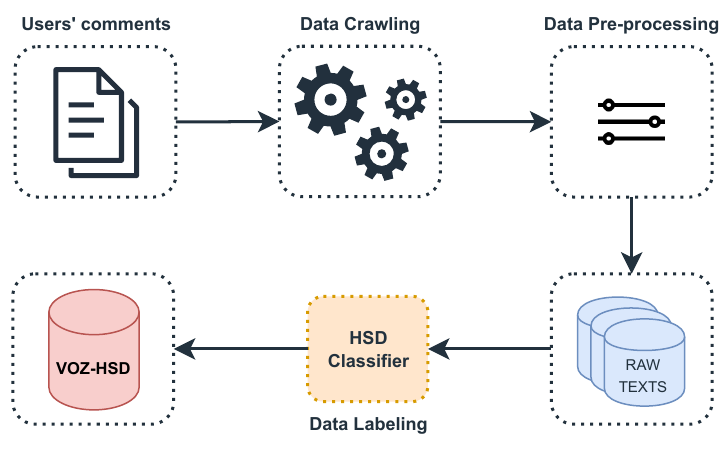}
    \caption{The process of creating VOZ-HSD by the automated data labeling approach.}
    \label{fig:automated-data-labeling}
\end{figure}

% Automated data creation
Vietnamese is a low-resource language, which results in a need for more extensive datasets for training targeted language models, particularly in specific NLP tasks. In this research, we present a massive Vietnamese hate speech classification dataset alongside an automated data annotation system. Figure \ref{fig:automated-data-labeling} illustrates the entire process, which includes several modules.

% Data crawling
\textbf{Data Crawling.} Initially, data was crawled from VOZ Forums\footnote{https://voz.vn/}, recognized as one of the most popular forums among young Vietnamese individuals. In comparison to other mainly used social media platforms like Facebook or TikTok, which have been utilized as pre-training data sources for other transfer learning models \cite{nguyen-etal-2023-visobert}, data sourced from VOZ presents a potentially richer resource due to its characteristic of unrestricted freedom of speech. Consequently, it represents a valuable asset for research into hate speech.

The primary source of data collection was the main chat parent-thread\footnote{https://voz.vn/\#khu-vui-choi-giai-tri.16}, where users typically share their personal thoughts, often incorporating toxic content and emotional expressions. The crawling process involved the utilization of the BeautifulSoup4\footnote{https://pypi.org/project/beautifulsoup4/} tool.

% Data pre-processing
\textbf{Data Pre-processing.} Given that the raw data comprises social media content, it includes noise and undesirable elements like user identities, URLs, or references to other comments. Therefore, pre-processing texts is exceedingly crucial before inputting them into models. In our research, we adopt the data pre-processing approach outlined by \citet{nguyen-etal-2023-visobert}, which involves tasks such as eliminating mentioned links, @username, retaining emojis and emoticons, and further excluding quotes, considered a distinctive element in a forum-based social media platform. The process results in approximately 1.7GB of uncompressed textual data. 

% Data annotator
\textbf{AI Data Annotator.} The advancements observed in AI data labeling systems \cite{desmond2021semi} have motivated our research to explore automated data annotation, to generate extensive datasets for hate speech classification. Since we experiment with pre-trained models using different data ratios that require raw texts to be labeled (as discussed in Section \ref{sec:discussion}), we initially convert the ViHSD dataset \cite{luu2021large}, a recognized benchmark for hate speech detection in Vietnamese, into two labels: {\fontfamily{pcr}\selectfont CLEAN} $\Rightarrow$ {\fontfamily{pcr}\selectfont NONE}, and ({\fontfamily{pcr}\selectfont OFFENSIVE}, {\fontfamily{pcr}\selectfont HATE}) $\Rightarrow$ {\fontfamily{pcr}\selectfont HATE}, and employ it as a training dataset for training our classifier.

Following this, we fine-tuned several pre-trained models designed for Vietnamese to identify the best-performing ones. Results, shown in Table \ref{tab:classifier-performance} in Appendix \ref{sec:appendix-fine-tune-classifier}, reveal that the ViSoBERT-based fine-tuned model achieves the highest Macro F1-score. Thus, we select this model as the HSD Classifier.

% Data labeling
\textbf{Automated Data Labeling.} Utilizing the selected HSD Classifier, we automatically label all textual data within the raw dataset. The resultant dataset comprises approximately 10 million user comments annotated with hate speech labels. According to the statistics, there are over 500K+ comments labeled as containing the harmful content, constituting a significant portion of the total dataset. Notably, the dataset still maintains a substantial number of comments labeled as hate speech, particularly when compared to label distributions observed in the previous study by \citet{luu2021large} focusing on the Vietnamese hate speech detection task. We designate the final dataset as VOZ-HSD, indicating its purpose for hate speech detection and its origin from VOZ.

\begin{figure}[H]
    \centering
    \includegraphics[width=0.5\textwidth]{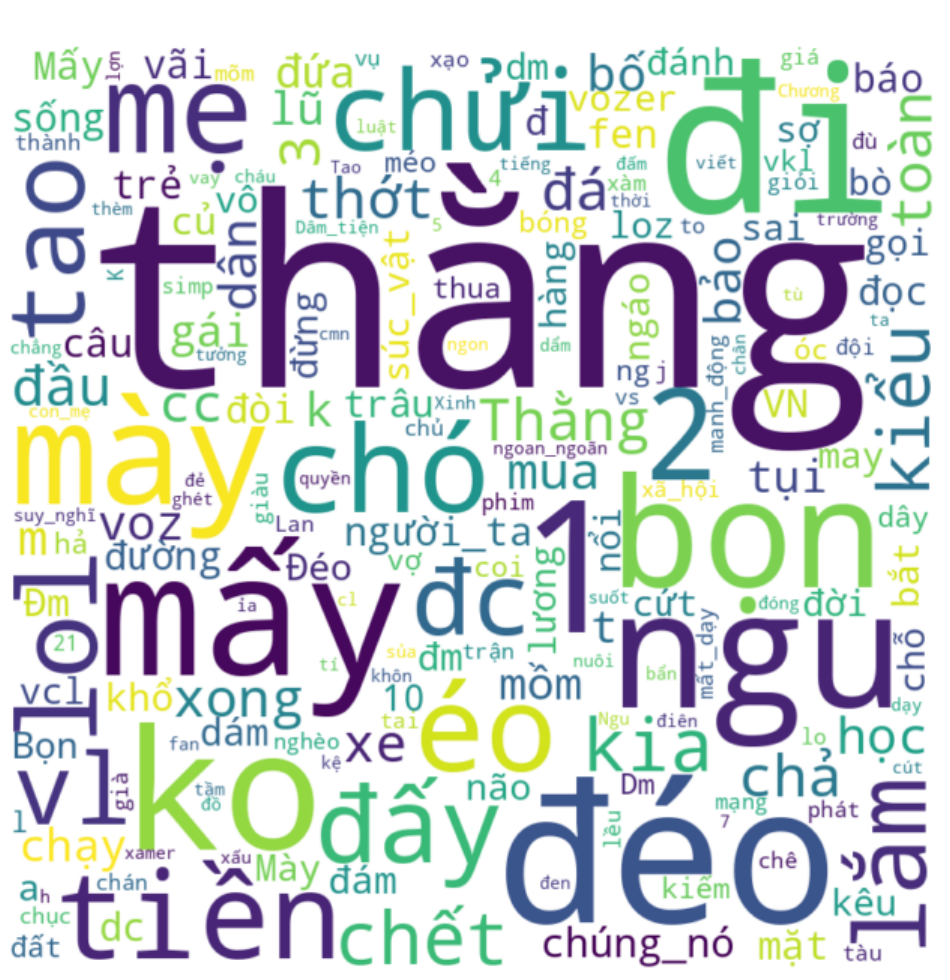} % setting width to half of text width
    \caption{The word cloud of VOZ-HSD dataset.}
    \label{fig:VOZ-HSD-wordcloud}
\end{figure}

The word cloud shown in Figure \ref{fig:VOZ-HSD-wordcloud} highlights the most common terms in harmful comments within the VOZ-HSD dataset. Notably, the dataset is rich in Vietnamese offensive language, featuring words like "thằng" (guy), "ngu" (stupid), "đéo" (fuck), and "chó" (dog). These terms encompass profanity, explicit content, colloquialisms, and informal expressions. Consequently, this dataset is valuable for enhancing the ability of language models to detect and mitigate hate speech effectively.

% About the process of the pre-training model
\subsection{Model Pre-training}
Domain-specific pre-training and the efficacy of text-to-text models in addressing HSD tasks \cite{ranasinghe2023text} have attained significant achievements. Following that, we embark on the pre-training of the \textsc{ViHateT5} model, leveraging the T5 architecture. The constructed VOZ-HSD dataset is employed as the pre-training dataset, comprising samples extracted from real-life comments.

% About the process of fine-tuning the model
\subsection{Model Fine-tuning}
To evaluate the trained model, we proceed to fine-tune the pre-trained \textsc{ViHateT5} on various hate-speech-based datasets currently available, focusing on three tasks in Vietnamese.

% About ViHSD
\textbf{Hate Speech Detection (ViHSD).} Initially devised to identify harmful content in user comments across social media platforms in Vietnam, the Vietnamese Hate Speech Detection (ViHSD) dataset \cite{luu2021large} has been extensively employed for text classification tasks. It involves categorizing texts into three labels: {\fontfamily{pcr}\selectfont HATE}, {\fontfamily{pcr}\selectfont OFFENSIVE}, and {\fontfamily{pcr}\selectfont CLEAN}. The ViHSD dataset comprises over 33K comments collected from comment sections of Facebook pages and YouTube videos.

% About ViCTSD
\textbf{Toxic Speech Detection (ViCTSD).} The ViCTSD dataset \cite{nguyen2021constructive} was initially formulated to identify constructiveness and toxicity in user comments. Originating from online news websites, where users are typically middle-aged individuals who often express themselves in formal styles, the level of offensiveness may not be as overt as in other datasets. Consequently, conducting toxicity detection on this dataset challenges language models.

% About ViHOS
\textbf{Hate Spans Detection (ViHOS).} The ViHOS dataset \cite{hoang-etal-2023-vihos} represents the first human-annotated corpus for identifying hateful and offensive spans within Vietnamese texts, providing a syllable-level task for HSD in Vietnamese. With over 11K comments and 26K annotated spans, this dataset diverges from its predecessors by focusing on the syllable level of hate speech, thereby augmenting the complexity of the task. Current BERT-based fine-tuning approaches typically adopt IOB sequence taggings to pre-process data, treating this task as a token classification task.

%% file: sections/4_experiments.tex
\section{Experiments and Results}
\label{sec:experiments}

\subsection{Data}
\begin{table*}[h]
\centering
\resizebox{\textwidth}{!}{%
\begin{tabular}{lcccll}
\hline
\multicolumn{1}{c|}{\textbf{Dataset}} &
  \multicolumn{3}{c|}{\textbf{Samples}} &
  \multicolumn{1}{c|}{\textbf{Labels}} &
  \multicolumn{1}{c}{\textbf{Source(s)}} \\ \hline
\multicolumn{6}{l}{\textbf{Pre-training Data}}                                                          \\ \hline
\multicolumn{1}{l|}{VOZ-HSD} & \multicolumn{3}{c|}{10.8M} & \multicolumn{1}{l|}{{\fontfamily{pcr}\selectfont NONE}, {\fontfamily{pcr}\selectfont HATE}} & Voz Forum \\ \hline
\multicolumn{6}{l}{\textbf{Finetuning Data}}                                                            \\ \hline
\multicolumn{1}{l|}{ViHSD (Binary)} &
  24,048 &
  2,672 &
  \multicolumn{1}{c|}{6,680} &
  \multicolumn{1}{l|}{{\fontfamily{pcr}\selectfont NONE}, {\fontfamily{pcr}\selectfont HATE}} &
  Facebook, Youtube \\
\multicolumn{1}{l|}{ViHSD \cite{luu2021large}} &
  24,048 &
  2,672 &
  \multicolumn{1}{c|}{6,680} &
  \multicolumn{1}{l|}{{\fontfamily{pcr}\selectfont CLEAN}, {\fontfamily{pcr}\selectfont OFFENSIVE}, {\fontfamily{pcr}\selectfont HATE}} &
  Facebook, Youtube \\
\multicolumn{1}{l|}{ViCTSD \cite{nguyen2021constructive}} &
  7,000 &
  2,000 &
  \multicolumn{1}{c|}{1,000} &
  \multicolumn{1}{l|}{{\fontfamily{pcr}\selectfont NONE}, {\fontfamily{pcr}\selectfont TOXIC}} &
  VnExpress \\
\multicolumn{1}{l|}{ViHOS \cite{hoang-etal-2023-vihos}} &
  8,844 &
  1,106 &
  \multicolumn{1}{c|}{1,106} &
  \multicolumn{1}{l|}{Hate Speech Spans} &
  Facebook, Youtube \\ \hline
\end{tabular}%
}
\caption{Statistics of datasets used in the experiments. Note that all samples in datasets are comments written in Vietnamese.}
\label{tab:data-statistics}
\end{table*}

% Overview
In this section, we outline the experiments conducted, including details regarding the training data utilized, model configurations, baselines, evaluation metrics, and the ensuing results. We illustrate our proposed \textsc{ViHateT5} model performance across multiple HSD tasks in Vietnamese, providing insights into its efficacy relative to other previous state-of-the-art approaches. Note that no specific data pre-processing is applied to any models to ensure fairness. 

% Table \ref{tab:data-statistics} presents a summary of the statistics for all datasets utilized in this study. The focus is on four distinct datasets: ViHSD \cite{luu2021large}, VLSP-HSD \cite{vu2020hsd}, ViCTSD \cite{nguyen2021constructive}, and ViHOS \cite{hoang-etal-2023-vihos}. These datasets are chosen for their pertinence to hate speech detection tasks in Vietnamese and are predominantly sourced from social media platforms. It is worth noting that we follow the original instructions for data splitting and employ their pre-processing techniques to ensure the robust performance of the models in this study.

% Pre-training data VOZ-HSD
\textbf{Pre-training Data.} The raw texts from the VOZ-HSD dataset are used for the pre-training phase of the \textsc{ViHateT5} model. As labeled by the HSD Classifier, raw texts in the VOZ-HSD dataset with generated labels can be helpful for data analysis and experiments with different proportions of hate-labeled samples.

\begin{table*}[ht]
\centering
\resizebox{\textwidth}{!}{%
\begin{tabular}{lccclcc}
\hline
\multicolumn{1}{c}{\textbf{Model}} &
  \textbf{\#archs} &
  \textbf{\#params} &
  \textbf{\#max\_len} &
  \multicolumn{1}{c}{\textbf{Data Domain}} &
  \textbf{\#vocab} &
  \textbf{Size} \\ \hline
BERT (multilingual, cased) \cite{devlin-etal-2019-bert}   & base  & 177M & 512 & BookCorpus+EnWiki         & 120K & 20GB  \\
BERT (multilingual, uncased) \cite{devlin-etal-2019-bert} & base  & 167M & 512 & BookCorpus+EnWiki         & 106K & 20GB  \\
DistilBERT (multilingual) \cite{sanh2019distilbert}       & base  & 135M & 512 & BookCorpus+EnWiki         & 120K & 20GB  \\
XLM-RoBERTa \cite{conneau2019cross}                       & base  & 270M & 512 & CommonCrawl               & 250K & 2.5TB \\ \hline
PhoBERT \cite{nguyen-tuan-nguyen-2020-phobert}            & base  & 135M & 256 & ViWiki+ViNews             & 64K  & 20GB  \\
PhoBERT\_v2 \cite{nguyen-tuan-nguyen-2020-phobert}        & base  & 135M & 256 & ViWiki+ViNews+OscarCorpus & 64K  & 140GB \\
viBERT \cite{tran2020improving}                           & base  & 115M & 256 & Vietnamese News           & 38K  & 10GB  \\
ViSoBERT \cite{nguyen-etal-2023-visobert}                 & base  & 98M  & 256 & Vietnamese Social Media   & 15K  & 1GB   \\ \hline
\textbf{\textsc{ViHateT5} (Ours)}                                           & base  & 223M    & 256   & VOZ-HSD                  & 32K    & 1.7GB \\ \hline
\end{tabular}%
}
\caption{Details on baseline pre-trained models and our \textsc{ViHateT5} used in the experiments, including model architecture, number of total parameters, max sequence length, pre-training data domain, vocab size, and the total data size. Note that the data size for pre-training multilingual models reflects the total, not just Vietnamese texts.}
\label{tab:model-information}
\end{table*}

% Downstream tasks
\textbf{Downstream Task Data.} Next, we select several benchmark datasets for hate speech detection in Vietnamese to assess the performance of our proposed \textsc{ViHateT5} model compared to others. These datasets encompass both sentence-level tasks, such as hate speech detection and toxic speech detection, and syllable-level tasks, such as hate spans detection, using the ViHSD, ViCTSD, and ViHOS datasets, respectively.

% Data pre-processing for T5
\textbf{Specific Data Pre-processing for T5-based Models.} Given the utilization of a text-to-text architecture, T5-based models require specific data pre-processing prior to fine-tuning downstream tasks. Figure \ref{fig:ViHateT5-example} illustrates the multitasking input-output of our proposed \textsc{ViHateT5} model, along with other T5-based models. Initially, we append task-specific prefixes, namely {\fontfamily{pcr}\selectfont `hate-speech-detection'}, {\fontfamily{pcr}\selectfont `toxic-speech-detection'}, and {\fontfamily{pcr}\selectfont `hate-spans-detection'} for texts sourced from the ViHSD, ViCTSD, and ViHOS datasets, respectively. For the syllable-level task of ViHOS, we incorporate tags {\fontfamily{pcr}\selectfont [HATE]} before and after the spans to include multiple spans based on the given index spans, thereby producing target texts for model training. Table \ref{tab:data_example} in Appendix \ref{sec:appendix-data-sample} provides several samples of processed texts for both BERT-based and T5-based models employed in this study.

% Model parameters
\subsection{Model Setup}
We follow the original pre-training strategy outlined for the T5 model \cite{raffel2023exploring} to pre-train our \textsc{ViHateT5}. Both training and validation are conducted with 128 batch size. The process for pre-training is executed over 20 epochs, employing the Adam optimizer with a lower learning rate set at 5e-3. Additionally, a weight decay of 0.001 is applied, with the initial 2,000 steps designated for warm-up during training.

In the fine-tuning phase, we maintain uniform settings for all BERT-based baseline models across specific tasks. Similarly, the same model settings are applied to T5-based models. For detailed information regarding the model settings for fine-tuning downstream tasks, please refer to Appendix \ref{sec:appendix-experimental-setting-fine-tuning}.

It is worth noting that all experiments are carried out with a limited resource setup utilizing a single NVIDIA A6000 GPU.

% Baseline BERT
\subsection{Baseline}
We establish various baselines based on BERT-based architecture to compare with the performance of our proposed \textsc{ViHateT5} model. The selected BERT-based pre-trained language models, encompassing both multilingual and monolingual variants, are readily available and extensively utilized for Vietnamese. Details of these pre-trained models, along with our proposed \textsc{ViHateT5}, are provided in Table \ref{tab:model-information}. This information includes their architectures, total parameters, maximum sequence length of the model, pre-training data domain, vocabulary size, and data size.

% Multilingual BERT
\subsubsection{Multilingual Pre-trained Models}
\textbf{BERT}. \citet{devlin-etal-2019-bert} introduced their remarkable pre-training language model called BERT, representing a landmark achievement in NLP tasks. Unlike previous models, BERT leverages the Transformer architecture for pre-training on extensive corpora, simultaneously leveraging both left and right context in every layer. This bidirectional representation grants BERT a contextual understanding of language surpassing prior methods. Consequently, BERT performed excellently on various NLP tasks. Furthermore, BERT only needs minimal fine-tuning for downstream tasks, demonstrating its remarkable generalizability. As a result, BERT has become a fundamental component in NLP pipelines. Besides the original English-only pre-trained model, the multilingual version supporting 100+ languages, including Vietnamese, has been released to solve problems in various languages. 

\textbf{DistilBERT}, introduced by \citet{sanh2019distilbert}, is a compact pre-trained language model derived from the popular BERT architecture. Knowledge distillation effectively captures the knowledge of a larger pre-trained model while drastically reducing its size and computational complexity. Despite its smaller size, DistilBERT exhibits remarkable performance, retaining over 95\% of BERT's accuracy on the GLUE language understanding benchmark. The multilingual version of DistilBERT has also been publicly released.
    
\textbf{XLM-RoBERTa}, developed by \citet{conneau2019cross}, stands as a prominent achievement in multilingual language modeling. Leveraging a Transformer architecture and trained on a massive dataset of 2.5TB filtered CommonCrawl text across 100 languages, it employs a masked language modeling (MLM) objective to learn cross-lingual representations. This approach surpasses previous models like mBERT in diverse tasks. As a result, XLM-R offers a compelling solution for cross-lingual tasks, pushing the boundaries of multilingual language understanding.

% Monolingual BERT
\subsubsection{Monolingual Pre-trained Models}
\textbf{PhoBERT}, known as one of the public large-scale language models for Vietnamese, has achieved excellent performance on various NLP tasks \cite{nguyen-tuan-nguyen-2020-phobert} since its emergence. PhoBERT initially comes in two sizes, base and large, catering to different needs and computing resources. Built upon the successful RoBERTa architecture, PhoBERT offers improved robustness and performance. PhoBERT is now available in three versions, including base, large, and base version-2nd.
    
\textbf{viBERT} \cite{tran2020improving} has been released, aiming at improving performance on Vietnamese language tasks. It is similar to the well-known BERT model but trained explicitly on a massive corpus of Vietnamese text data. The architecture of the viBERT base version is with 12 transformer blocks and 768 hidden units. 
    
\textbf{ViSoBERT.} As known as the current state-of-the-art model, which was trained on social media data, ViSoBERT outperformed previous models in terms of social media NLP tasks \cite{nguyen-etal-2023-visobert}. The architecture of ViSoBERT is based on XLM-R with 12 encoder layers and 768 hidden dimensions. Because of its domain-specific pre-training, ViSoBERT is able to tokenize unusual and spoken-like content so that it can surpass existing models in various tasks.

% Evaluation
\subsection{Evaluation}
The downstream tasks in this study are evaluated using metrics consistent with those employed in previous publications \cite{nguyen-etal-2022-smtce,nguyen-etal-2023-visobert}, which include accuracy score (Acc), weighted F1-score (WF1), and macro F1-score (MF1). For each task, MF1 serves as the primary evaluation metric, as per the original research. Besides, we calculate the Average MF1, derived from the MF1 scores across three benchmark datasets, to depict the overall performance of each model on HSD tasks.

Additionally, hate spans detection is a syllable-level task, necessitating the processing of output from T5-based models before computing evaluation metrics. To accomplish this, we follow Process \ref{pro:t5-input-process} to obtain index spans consistent with the original dataset structure.

\vspace{3pt}
\begin{algorithm}[h]
  \caption{Index spans retrieval from T5-based models' output}\label{pro:t5-input-process}
  
  \KwData{{\fontfamily{pcr}\selectfont [HATE]}vcl{\fontfamily{pcr}\selectfont [HATE]} thật. Chịu luôn {\fontfamily{pcr}\selectfont [HATE]}đm m{\fontfamily{pcr}\selectfont [HATE]}!!! \\(Original text $\mathbb{T}$: ``vcl thật. Chịu luôn đm m!!!")}
  
  \KwResult{[0, 1, 2, 20, 21, 22, 23]}
  
  \BlankLine
  Construct list $\mathbb{H}$ containing sub-strings covered by two {\fontfamily{pcr}\selectfont [HATE]} tokens\;
  
  \BlankLine
  From $\mathbb{H}$, find the corresponding index spans $\mathbb{I}$ of each sub-string in the original text $\mathbb{T}$\;
  
  \BlankLine
  \Return{$\mathbb{I}$}\;

\end{algorithm}
\vspace{3pt}

Next, we construct the binary form of indices by Process \ref{pro:binary-presentation}. Also, note that this second process is also applied to the ground truth data in order to compute the evaluation metrics consistently.

\vspace{3pt}
\begin{algorithm}[h]
  \caption{Converting index spans for evaluation computation}\label{pro:binary-presentation}
  
  \KwData{[0, 1, 2, 20, 21, 22, 23] (along with the original text $\mathbb{T}$)}
  
  \KwResult{[1, 1, 2, 0, 0, 0, 0, 0, 0, 0, 0, 0, 0, 0, 0, 0, 0, 0, 0, 0, 1, 1, 1, 1, 0, 0, 0]}
  
  \BlankLine
  Calculate the length $\mathbb{L}$ of the original text $\mathbb{T}$\;

  \BlankLine
  Initialize the list $\mathbb{F}$ with `0' elements corresponding to the length $\mathbb{L}$\;
  
  \BlankLine
  Replace elements in list $\mathbb{F}$ whose indices are in list index spans $\mathbb{I}$ to `1'\;

  \BlankLine
  \Return{$\mathbb{F}$}\;
\end{algorithm}
\vspace{3pt}

% Results
\subsection{Experimental Results}

Table \ref{tab:experimental-results} shows the performance of \textsc{ViHateT5} compared to other approaches across various HSD tasks. Through experiments conducted under identical settings, \textsc{ViHateT5} consistently outperforms other models, establishing itself as SOTA for most HSD-related tasks in Vietnamese.

\begin{table*}[]
\centering
\resizebox{\textwidth}{!}{%
\begin{tabular}{l|c|ccc|ccc|ccc}
\hline
\multicolumn{1}{c|}{\multirow{2}{*}{\textbf{Model}}} &
  \multicolumn{1}{l|}{\multirow{2}{*}{\textbf{Average MF1}}} &
  \multicolumn{3}{c|}{\textbf{Hate Speech Detection}} &
  \multicolumn{3}{c|}{\textbf{Toxic Speech Detection}} &
  \multicolumn{3}{c}{\textbf{Hate Spans Detection}} \\ \cline{3-11} 
\multicolumn{1}{c|}{} &
  \multicolumn{1}{l|}{} &
  \textbf{\textbf{Acc}} &
  \textbf{\textbf{WF1}} &
  \textbf{\textbf{MF1}} &
  \textbf{\textbf{Acc}} &
  \textbf{\textbf{WF1}} &
  \textbf{\textbf{MF1}} &
  \textbf{\textbf{Acc}} &
  \textbf{\textbf{WF1}} &
  \textbf{\textbf{MF1}} \\ \hline
BERT (multilingual, cased)   & 0.6930 & 0.8736 & 0.8680 & 0.6444 & 0.8983 & 0.8855 & 0.6710 & 0.8601 & 0.8464 & 0.7637 \\
BERT (multilingual, uncased) & 0.6827 & 0.8666 & 0.8606 & 0.6292 & 0.8993 & 0.8877 & 0.6796 & 0.8520 & 0.8172 & 0.7393 \\
DistilBERT (multilingual)    & 0.6933 & 0.8630 & 0.8606 & 0.6334 & 0.8962 & 0.8873 & 0.6850 & 0.8585 & 0.8428 & 0.7615 \\
XLM-RoBERTa                  & 0.7265 & 0.8729 & 0.8697 & 0.6508 & 0.9015 & 0.9007 & 0.7153 & 0.8834 & 0.8754 & 0.8133 \\ \hline
PhoBERT                      & 0.6963 & 0.8675 & 0.8652 & 0.6476 & 0.9078 & 0.9027 & 0.7131 & 0.8465 & 0.8112 & 0.7281 \\
PhoBERT\_v2                  & 0.7050 & 0.8742 & 0.8733 & 0.6660 & 0.9023 & 0.8978 & 0.7139 & 0.8492 & 0.8151 & 0.7351 \\
viBERT                       & 0.6780 & 0.8633 & 0.8579 & 0.6285 & 0.8881 & 0.8817 & 0.6765 & 0.8463 & 0.8128 & 0.7291 \\
ViSoBERT                     & 0.7507 & 0.8817 & 0.8786 & 0.6771 & 0.9035 & 0.9016 & 0.7145 & 0.9016 & 0.9007 & 0.8604 \\ \hline
\textbf{\textsc{ViHateT5} (Ours)} &
  \textbf{0.7556} &
  \textbf{0.8876} &
  \textbf{0.8914} &
  \textbf{0.6867} &
  \textbf{0.9080} &
  \textbf{0.9178} &
  \textbf{0.7163} &
  \textbf{0.9100} &
  \textbf{0.9020} &
  \textbf{0.8637} \\ \hline
\end{tabular}%
}
\caption{Comparative performance results of diverse models, including fine-tuned models from multilingual pre-trained language models, monolingual models, and our proposed \textsc{ViHateT5} model. Evaluation metrics include Accuracy (Acc), Weighted F1-score (WF1), and Macro F1-score (MF1) across various hate speech detection (HSD)-related tasks.}
\label{tab:experimental-results}
\end{table*}

In the realm of sentence-level tasks, specifically hate speech detection on the ViHSD dataset and toxic speech detection on the ViCTSD dataset, our proposed \textsc{ViHateT5} model demonstrates outstanding performance, surpassing previous models with MF1 scores of 68.67\% and 71.63\%, respectively. Meanwhile, for the remaining baseline models, ViSoBERT achieves its highest performance on the hate speech detection task with an MF1 score of 67.71\%, whereas XLM-RoBERTa attains the highest MF1 score of 71.53\% for toxic speech detection.

In the domain of syllable-level tasks, such as hate spans detection, the \textsc{ViHateT5} model showcases its effective ability to identify harmful segments by leveraging its text-to-text architecture, achieving the highest MF1 score of 86.37\%. Additionally, ViSoBERT ranks second on the leaderboard with an MF1 score of 86.04\%. Both models, being pre-trained specifically on social media domain data, yield consistent results on social media benchmark datasets, with a relatively small gap between them\footnote{The performance of ViSoBERT reported by \citet{nguyen-etal-2023-visobert} on similar tasks may slightly differ from our experiments due to variations in model settings during reproduction owing to resource constraints.}.

% Analysis
\subsection{Discussion}
\label{sec:discussion}
In this section, we delve into the comparison between the unified HSD-fine-tuned \textsc{ViHateT5} model and other T5-based models fine-tuned on HSD-related tasks in Vietnamese. Additionally, we explore the performance of our proposed pre-trained \textsc{ViHateT5} model across different pre-training data settings. Furthermore, we assess the model's ability to tackle syllable-level tasks.

% ViHateT5 vs other T5
\textbf{\textsc{ViHateT5} vs. other T5-based models.} The effectiveness of the T5 text-to-text architecture in addressing HSD tasks in Vietnamese has been demonstrated by \textsc{ViHateT5}. This study evaluates other T5-based models supporting Vietnamese for HSD tasks. We experiment with mT5-base, mT5-large \cite{xue2021mt5} for multilingual models, and ViT5-base, ViT5-large \cite{phan2022vit5} for monolingual models. The fine-tuning phases of these models are conducted under the same settings as \textsc{ViHateT5}, listed in Table \ref{tab:T5-based-parameters} in Appendix \ref{sec:appendix-experimental-setting-fine-tuning-T5}. Due to resource limitations, the batch size for large versions is reduced. Table \ref{tab:T5-based-performance} compares the performance of \textsc{ViHateT5} with other T5-based models across three benchmark HSD datasets in Vietnamese.

\begin{table}[H]
\centering
\resizebox{\columnwidth}{!}{%
\begin{tabular}{cccccc}
\hline
\textbf{\textbf{\textbf{\textbf{}}}} &
  \textbf{\textbf{\#archs}} &
  \textbf{\textbf{\textbf{\textbf{\textbf{ViHSD}}}}} &
  \textbf{\textbf{\textbf{\textbf{\textbf{ViCTSD}}}}} &
  \textbf{\textbf{\textbf{\textbf{\textbf{ViHOS}}}}} &
  \textbf{Average} \\ \hline
mT5      & base & 0.6676          & 0.6993          & 0.8660          & 0.7289          \\
ViT5     & base & 0.6695          & 0.6482          & \textbf{0.8690} & 0.7443          \\ \hline
\textbf{\textsc{ViHateT5}} & base & \textbf{0.6867} & \textbf{0.7163} & 0.8637          & \textbf{0.7556} \\ \hline
\end{tabular}%
}
\caption{\textsc{ViHateT5} versus other T5-based models regarding Vietnamese HSD-related task performance with Macro F1-score.}
\label{tab:T5-based-performance}
\end{table}
The results attained highlight the superior performance of our proposed \textsc{ViHateT5} model across various HSD-related tasks in comparison to other T5-based models supporting Vietnamese. The primary reason for this disparity lies in the nature of HSD benchmark datasets, which predominantly consist of spoken textual data, such as users' comments on the internet. These data exemplify social media characteristics, comprising informal written style texts accompanied by abbreviations, emojis, or teencode.

In contrast, while mT5 and ViT5 were pre-trained on formal content sources such as news or wiki pages, \textsc{ViHateT5} was pre-trained on a domain-specific social media pre-training dataset. This domain-specific pre-training dataset ensures that \textsc{ViHateT5} is more adept at understanding and processing informal language used in social media contexts, thereby yielding superior performance on HSD tasks.

% Data Ratio Pre-training
\textbf{How Pre-training Data Affects \textsc{ViHateT5}.} The effectiveness of pre-training a transformer model on a domain-specific dataset was further validated by the ViSoBERT model, which demonstrated superior performance across various social media benchmark datasets \cite{nguyen-etal-2023-visobert}. In this work, we assess how varying the data ratio in pre-training data affects our proposed models. This evaluation involves pre-training under different data conditions: utilizing full-data samples as in this study, employing a balanced-label pre-trained model with equal samples for both labels and utilizing a hate-only pre-trained model where only hate labels are retained for pre-training.

Based on the generated labels, we conducted experiments to pre-train \textsc{ViHateT5} under different data ratio conditions. The first condition used the entire dataset, while the second balanced the labels by reducing the number of {\fontfamily{pcr}\selectfont CLEAN} samples. The final condition exclusively pre-trained on {\fontfamily{pcr}\selectfont HATE} labeled samples. Table \ref{tab:different-pre-training-data} presents the performance of these models after fine-tuning them on downstream tasks. It is worth noting that due to the relatively small size of the training samples in the 100\% ratio condition, which is not sufficient for pre-training from scratch, we opted to use the continual pre-training approach for all these experiments, utilizing weights from the ViT5-base\footnote{https://huggingface.co/VietAI/vit5-base}.

\begin{table}[H]
\centering
\resizebox{\columnwidth}{!}{%
\begin{tabular}{cccccc}
\hline
\textbf{\textbf{\textbf{Ratio}}} &
  \textbf{\textbf{\textbf{Samples}}} &
  \textbf{\textbf{Epochs}} &
  \textbf{\textbf{ViHSD}} &
  \textbf{\textbf{ViCTSD}} &
  \textbf{\textbf{ViHOS}} \\ \hline
\multirow{2}{*}{100\%}  & \multirow{2}{*}{584,495}    & 10 & 0.6548          & 0.6134          & 0.8542          \\
                        &                             & 20 & 0.6577          & 0.6258          & 0.8601          \\ \hline
\multirow{2}{*}{50\%}   & \multirow{2}{*}{1,168,990}  & 10 & 0.6600          & 0.6022          & 0.8577          \\
                        &                             & 20 & 0.6620          & 0.6642          & 0.8588          \\ \hline
\multirow{2}{*}{5.54\%} & \multirow{2}{*}{10,747,733} & 10 & 0.6286          & \textbf{0.7358} & 0.8591          \\
                        &                             & 20 & \textbf{0.6800} & 0.7027          & \textbf{0.8644} \\ \hline
\end{tabular}%
}
\caption{The performance of \textsc{ViHateT5}, measured by Macro F1-score, under various data pre-training conditions. The "Ratio" column indicates the percentage of hate data in the total dataset.}
\label{tab:different-pre-training-data}
\end{table}

The analysis reveals that pre-training with balanced or hate-labeled datasets does not improve model performance and can even lower MF1. However, different pre-training conditions affect \textsc{ViHateT5} performance across various HSD tasks, suggesting additional pre-training on another T5 architecture model could be beneficial despite limited data. Also, increasing the number of pre-training epochs improves performance. Further research could explore resource-intensive setups to enhance \textsc{ViHateT5} performance.

% Syllable-level ability
\textbf{ViHateT5 in Syllable-level Hate Speech Detection.} \textsc{ViHateT5} has demonstrated its effectiveness in tackling syllable-level challenges, particularly in detecting hate speech spans within the ViHOS dataset. Leveraging an innovative architecture and training methodology derived from the T5 text-to-text transformer architecture, \textsc{ViHateT5} surpasses baseline methods relying on BERT-based models, which primarily encounter limitations due to their token-level processing approach. Operating at the syllable level empowers \textsc{ViHateT5} to pinpoint harmful spans within textual contexts accurately. Furthermore, its text-to-text framework presents \textsc{ViHateT5} with opportunities to extend its capabilities to other tasks, such as hate speech detection question-answering or summarization, through adjustments to the prefix for fine-tuning.

%% file: sections/5_conclusion.tex
\section{Conclusions}
\label{sec:conclusion}

Advancements in hate speech detection tasks in Vietnamese have recently gained notable progress thanks to the use of transformer models. However, these efforts remain fragmented due to the reliance on separate fine-tuned models for distinct tasks. Hence, our research aims to introduce a unified text-to-text transformer model, \textsc{ViHateT5}, with the potential to address prevailing issues in hate speech detection in Vietnamese and attain state-of-the-art performance. Moreover, \textsc{ViHateT5}'s pre-training on domain-specific datasets enables it to grasp the nuances of social media content in Vietnamese deeply. The open-source nature of both the dataset and the model facilitates researchers and developers in leveraging our work, fostering further advancements in Vietnamese NLP and online safety.

%% file: sections/6_limitations.tex
\section{Limitations}
\label{sec:limitation}

Training a language model through pre-training demands a substantial volume of data and computational power. In this investigation, we built an initial pre-training dataset, VOZ-HSD, suitable for experimentation with the ViT5-base, based on the base version of T5 architecture. However, it might not be adequate for larger versions. Previous research \cite{phan2022vit5} outlines the effectiveness of these large-setting models, demonstrating their performance relative to smaller versions like the T5-base, which is employed in this experiment.

%% file: sections/7_ethical.tex
\section{Ethical Statements}
\label{sec:ethical}

% About VOZ-HSD data
The proposed \textsc{ViHateT5} model is specifically designed to handle various hate speech detection tasks in the Vietnamese language. Trained on a substantial auto-labeled dataset VOZ-HSD, as discussed in Section \ref{sec:automated-data-creation}, the collected data undergoes meticulous preprocessing to eliminate all user identities, safeguarding user privacy.

% ViHateT5 contributions to social
With the rise of social media platforms and the corresponding increase in harmful content, there are unintended repercussions that require content moderation to protect users in online conversations. The proposed \textsc{ViHateT5} model aims to make a meaningful contribution by delivering accurate performance across various hate speech detection tasks in Vietnamese. This initiative seeks to enhance content moderation on social media, promoting transparency and fostering a healthier online environment.

%% file: sections/8_acknowledgement.tex
\section*{Acknowledgement}

This research was supported by The VNUHCM-University of Information Technology's Scientific Research Support Fund. We would like to express our gratitude to the anonymous ACL reviewers for their valuable and constructive feedback. Their contributions have significantly enriched the quality and thoroughness of our work.

%% file: sections/9_appendix.tex
\appendix
\onecolumn

% HSD Classifier
\section{Fine-tuning Hate Speech Classifier}
\label{sec:appendix-fine-tune-classifier}
To develop hate speech classifiers, we fine-tune existing pre-trained language models designed for the Vietnamese language. All experiments utilize a common set of pre-training language models. The training process is conducted over 3 epochs, employing a batch size of 16 for both training and evaluation phases. The maximum sequence length is defined as 128, and the learning rate is set to 1e-5. The remaining parameters adhere to recommendations from prior research. Table \ref{tab:classifier-performance} provides an overview of the performance of various models in detecting hate speech in Vietnamese, utilizing two labels: {\fontfamily{pcr}\selectfont HATE} and {\fontfamily{pcr}\selectfont NONE}. Note that selecting the best model for our proposed system is based on its classification performance using the Macro F1-score (MF1). The chosen hate speech classifier, ViSoBERT-HSD, is publicly available at HuggingFace\footnote{https://huggingface.co/tarudesu/ViSoBERT-HSD}. 

\begin{table*}[h]
\centering
\resizebox{\textwidth}{!}{%
\begin{tabular}{lccc}
\hline
\multicolumn{1}{c}{\textbf{}}                                   & \textbf{Accuracy} & \textbf{Weighted F1-score} & \textbf{Macro F1-score} \\ \hline
\multicolumn{4}{l}{\textbf{Multilingual Pre-trained Models}}                               \\ \hline
BERT (Multilingual, base, cased) \cite{devlin-etal-2019-bert}   & 0.8615            & 0.8483                     & 0.7089                  \\
BERT (Multilingual, base, uncased) \cite{devlin-etal-2019-bert} & 0.8524            & 0.8335                     & 0.6742                  \\
DistilBERT (Multilingual) \cite{sanh2019distilbert}    & 0.8344 & 0.7992 & 0.5895          \\
XLM-RoBERTa (base) \cite{conneau2019cross}             & 0.8477 & 0.8070 & 0.5965          \\
XLM-RoBERTa (large) \cite{conneau2019cross}            & \textbf{0.8877} & 0.8836 & 0.7861          \\ \hline
\multicolumn{4}{l}{\textbf{Monolingual Pre-trained Models}}                                \\ \hline
PhoBERT (base) \cite{nguyen-tuan-nguyen-2020-phobert}  & 0.8603 & 0.8479 & 0.7095          \\
PhoBERT (large) \cite{nguyen-tuan-nguyen-2020-phobert} & 0.8678 & 0.8464 & 0.6936          \\
PhoBERT\_v2 (base) \cite{nguyen-tuan-nguyen-2020-phobert}       & 0.8754            & 0.8723                     & 0.7676                  \\
viBERT \cite{tran2020improving}                        & 0.8612 & 0.8463 & 0.7028          \\
ViSoBERT \cite{nguyen-etal-2023-visobert}              & 0.8477 & \textbf{0.9033} & \textbf{0.8227} \\ \hline
\end{tabular}%
}
\caption{Classification performances of various classifiers fine-tuned from different pre-trained models on the task of hate speech classification in order to find the best one for automated data annotation.}
\label{tab:classifier-performance}
\end{table*}

% Experiments
\section{Experimental Settings}
\label{sec:appendix-experimental-setting}

% Pre-training
\subsection{Model Pre-training}
\label{sec:appendix-experimental-setting-pre-training}
We initially pre-train \textsc{ViHateT5} from scratch on the VOZ-HSD dataset and its variants with different pre-training data settings with parameters illustrated in Table \ref{tab:pre-training-parameters}. Note that validation split means the ratio for the validation set taken from the original dataset. 

\begin{table}[H]
\centering
\resizebox{\textwidth}{!}{%
\begin{tabular}{ccccccccc}
\hline
\textbf{Name} &
  \textbf{Initial Weights} &
  \textbf{\#archs} &
  \textbf{Pre-training Data} &
  \textbf{Valid Split} &
  \multicolumn{1}{l}{\textbf{Epochs}} &
  \multicolumn{1}{l}{\textbf{l\_r}} &
  \multicolumn{1}{l}{\textbf{batch\_size}} &
  \multicolumn{1}{l}{\textbf{max\_seq\_len}} \\ \hline
\textsc{ViHateT5} & From scratch & base & VOZ-HSD                & 0.02 & 20       & 5e-3 & 128 & 256 \\
\textsc{ViHateT5} & ViT5-base    & base & VOZ-HSD                & 0.02 & [10, 20] & 5e-3 & 128 & 256 \\
\textsc{ViHateT5} & ViT5-base    & base & Balanced-label VOZ-HSD & 0.05 & [10, 20] & 5e-3 & 128 & 256 \\
\textsc{ViHateT5} & ViT5-base    & base & Hate-label VOZ-HSD     & 0.1  & [10, 20] & 5e-3 & 128 & 256 \\ \hline
\end{tabular}%
}
\caption{Model settings for pre-training \textsc{ViHateT5} variants. Note that all pre-trained models were trained on a limited-resource setting with a single GPU NVIDIA A6000.}
\label{tab:pre-training-parameters}
\end{table}

% Fine-tuning
\subsection{Model Fine-tuning}
\label{sec:appendix-experimental-setting-fine-tuning}

% BERT
\subsubsection{BERT-based Models}
To establish BERT-based models as baselines for fine-tuning each dataset, we implement the experimental configurations outlined below, as depicted in Table \ref{tab:BERT-based-parameters}. These settings adhere closely to those recommended in the original publications.

\begin{table}[H]
\centering
{%
\begin{tabular}{lccccc}
\hline
\multicolumn{1}{c}{\textbf{Dataset}} &
  \multicolumn{1}{l}{\textbf{batch\_size}} &
  \multicolumn{1}{l}{\textbf{max\_seq\_len}} &
  \multicolumn{1}{l}{\textbf{learning\_rate}} &
  \multicolumn{1}{l}{\textbf{weight\_decay}} &
  \multicolumn{1}{l}{\textbf{epochs}} \\ \hline
ViHSD   & 16 & 256 & 2e-5 & 0.01 & 4   \\
ViCTSD & 16 & 256 & 2e-5 & 0.01 & 4   \\
ViHOS  & 16 & 256  & 2e-5 & 0.01 & 10 \\ \hline
\end{tabular}%
}
\caption{Fine-tuning parameters for BERT-based models on each HSD-related task.}
\label{tab:BERT-based-parameters}
\end{table}

% T5
\subsubsection{T5-based Models}
\label{sec:appendix-experimental-setting-fine-tuning-T5}
The model configurations for fine-tuning T5-based models, including our \textsc{ViHateT5} utilized in this paper, are showcased in Table \ref{tab:T5-based-parameters}. The difference in the value of batch size occurs because of the limitation of GPU resources, leading to reduce the batch size for the training phase.

\begin{table}[H]
\centering
{%
\begin{tabular}{lccccc}
\hline
\multicolumn{1}{c}{\textbf{}} & \textbf{\#archs} & \textbf{batch\_size} & \textbf{max\_seq\_len} & \textbf{learning\_rate} & \textbf{epochs} \\ \hline
mT5            & base & 16 & 256 & 3e-4 & 4 \\
ViT5           & base & 32 & 256 & 3e-4 & 4 \\
\textsc{ViHateT5}-based & base & 32 & 256 & 3e-4 & 4 \\ \hline
\end{tabular}%
}
\caption{Fine-tuning parameters for T5-based models on the tasks of hate speech detection in Vietnamese. Note that \textsc{ViHateT5}-based indicates fine-tuned models from any variants of the pre-trained \textsc{ViHateT5}.}
\label{tab:T5-based-parameters}
\end{table}

% Data categories
\section{What is inside the VOZ-HSD dataset?}
\label{sec:appendix-VOZ-HSD}

Table \ref{tab:VOZ-HSD-distribution} illustrates the distribution of topic text data within the VOZ-HSD dataset. It is evident that we have gathered a wide range of conversation topics, indicating that the dataset is not skewed towards any particular domain and closely reflects real-life textual content. Previous studies by \citet{nguyen-etal-2023-visobert} have further demonstrated that even with a limited dataset size of only 1GB in an uncompressed format for pre-training a transformer on social media texts, the model can still exhibit strong performance across multiple tasks, achieving state-of-the-art results.

\begin{table}[H]
\centering
{%
\begin{tabular}{ccccc}
\hline
\textbf{No.} & \textbf{Parent Thread} & \textbf{N.o. Threads} & \textbf{N.o. Comments} & \textbf{Size (Uncompressed)} \\ \hline
1 & Random conversation    & 142,387 & 6,104,792 & 945MB \\
2 & News                   & 76,107  & 2,030,315 & 304MB \\
3 & Sports                 & 10,121  & 1,154,658 & 144MB \\
4 & Cars                   & 12,348  & 552,717   & 96MB  \\
5 & Movies - Music - Books & 6,467   & 329,601   & 52MB  \\
6 & Bikes                  & 5,093   & 258,728   & 41MB  \\
7 & Fashion                & 1,845   & 137,548   & 19MB  \\
8 & Food - Travel          & 3,492   & 136,649   & 19MB  \\
9 & Other hobbies          & 690     & 42,737    & 6MB   \\ \hline
\textbf{}    & \textbf{Total}         & \textbf{258,550}      & \textbf{10,747,745}    & \textbf{1.66GB}              \\ \hline
\end{tabular}
}
\caption{The distribution of comments in terms of conversation topics in the VOZ-HSD datasets. }
\label{tab:VOZ-HSD-distribution}
\end{table}

% Samples in the dataset
\section{Actual examples in benchmarks dataset and their pre-processed representations for BERT-based baseline models and our proposed \textsc{ViHateT5}}
\label{sec:appendix-data-sample}

\begin{landscape}
\begin{table}[h]
\centering
\resizebox{\columnwidth}{!}{%
\begin{tabular}{lclclc}
\hline
\multicolumn{2}{c}{\textbf{Examples}} &
  \multicolumn{2}{c}{\textbf{BERT-based Models}} &
  \multicolumn{2}{c}{\textbf{T5-based Models}} \\ \hline
\multicolumn{1}{c}{\textbf{Input}} &
  \textbf{Output} &
  \multicolumn{1}{c}{\textbf{Source}} &
  \textbf{Target} &
  \multicolumn{1}{c}{\textbf{Source}} &
  \textbf{Target} \\ \hline
\multicolumn{2}{c}{ViHSD} &
  \multicolumn{1}{c}{Original text} &
  \begin{tabular}[c]{@{}c@{}}{\fontfamily{pcr}\selectfont CLEAN} (0)\\ {\fontfamily{pcr}\selectfont OFFENSIVE} (1)\\ {\fontfamily{pcr}\selectfont HATE} (2)\end{tabular} &
  \multicolumn{1}{c}{Text with specific prefix} &
  \begin{tabular}[c]{@{}c@{}}{\fontfamily{pcr}\selectfont CLEAN}\\ {\fontfamily{pcr}\selectfont OFFENSIVE}\\ {\fontfamily{pcr}\selectfont HATE} \end{tabular} \\ \hline
\begin{tabular}[c]{@{}l@{}}Từ lý thuyết đến thực hành là cả 1 câu chuyện\\dài =))\\ (\textit{Translated: From theory to practice is a} \\ \textit{whole long story} =)))\end{tabular} &
  {\fontfamily{pcr}\selectfont CLEAN} &
  \begin{tabular}[c]{@{}l@{}}Từ lý thuyết đến thực hành\\ là cả 1 câu chuyện dài =))\end{tabular} &
  0 &
  \begin{tabular}[c]{@{}l@{}}hate-speech-detection: Từ lý thuyết\\ đến thực hành là cả 1 câu\\ chuyện dài =))\end{tabular} &
  {\fontfamily{pcr}\selectfont CLEAN} \\
\begin{tabular}[c]{@{}l@{}}Giống nhau như 2 giọt nước. Mà mỗi cái\\là 1 giọt nước mắt với 1 giọt nước sh!t thôi ạ\\ (\textit{Translated: Similar as two drops of water.}\\ \textit{But each one is a teardrop with a drop} \\ \textit{of shit too})\end{tabular} &
  {\fontfamily{pcr}\selectfont OFFENSIVE} &
  \begin{tabular}[c]{@{}l@{}}Giống nhau như 2 giọt nước.\\ Mà mỗi cái là 1 giọt nước mắt\\ với 1 giọt nước sh!t thôi ạ\end{tabular} &
  1 &
  \begin{tabular}[c]{@{}l@{}}hate-speech-detection: Giống nhau\\ như 2 giọt nước. Mà mỗi cái là 1\\ giọt nước mắt với 1 giọt nước sh!t \\thôi ạ\end{tabular} &
  {\fontfamily{pcr}\selectfont OFFENSIVE} \\
\begin{tabular}[c]{@{}l@{}}Im mẹ đi thằng mặt lon\\ (\textit{Translated: Shut up you big-faced idiot})\end{tabular} &
  {\fontfamily{pcr}\selectfont HATE} &
  Im mẹ đi thằng mặt lon &
  2 &
  \begin{tabular}[c]{@{}l@{}}hate-speech-detection: Im mẹ đi\\ thằng mặt lon\end{tabular} &
  {\fontfamily{pcr}\selectfont HATE} \\ \hline
\multicolumn{2}{c}{ViCTSD} &
  Original text &
  \begin{tabular}[c]{@{}c@{}}{\fontfamily{pcr}\selectfont NONE} (0)\\ {\fontfamily{pcr}\selectfont {\fontfamily{pcr}\selectfont TOXIC}} (1)\end{tabular} &
  \multicolumn{1}{c}{Text with specific prefix} &
  \begin{tabular}[c]{@{}c@{}}{\fontfamily{pcr}\selectfont NONE}\\ {\fontfamily{pcr}\selectfont TOXIC}\end{tabular} \\ \hline
\begin{tabular}[c]{@{}l@{}}Một thời để nhớ, bao kỷ niệm\\ gắn liền với những ca khúc của anh.\\ (\textit{Translated: A time to remember, with so many}\\ \textit{memories attached to his songs.})\end{tabular} &
  {\fontfamily{pcr}\selectfont NONE} &
  \begin{tabular}[c]{@{}l@{}}Một thời để nhớ, bao kỷ niệm\\ gắn liền với những ca khúc của anh.\end{tabular} &
  0 &
  \begin{tabular}[c]{@{}l@{}}
  toxic-speech-detecion: Một thời\\ để nhớ, bao kỷ niệm gắn liền với\\ những ca khúc của anh.\end{tabular} &
  {\fontfamily{pcr}\selectfont NONE} \\
\begin{tabular}[c]{@{}l@{}}nghe xong máu điên trong người\\ nổi lên. muốn đánh cho thằng cha một trận quá.....\\ (\textit{Translated: After listening, rage surged through}\\ \textit{my veins. I feel like giving that guy} \textit{a beating.....})\end{tabular} &
  {\fontfamily{pcr}\selectfont TOXIC} &
  \begin{tabular}[c]{@{}l@{}}nghe xong máu điên trong người\\ nổi lên. muốn đánh cho thằng cha\\ một trận quá.....\end{tabular} &
  1 &
  \begin{tabular}[c]{@{}l@{}}
  toxic-speech-detecion: nghe xong\\ máu điên trong người nổi lên.\\ muốn đánh cho thằng cha một\\ trận quá.....\end{tabular} &
  {\fontfamily{pcr}\selectfont TOXIC} \\ \hline
\multicolumn{2}{c}{ViHOS} &
  \multicolumn{1}{c}{Original text} &
  IOB Tags: {\fontfamily{pcr}\selectfont O, B-T, I-T} &
  \multicolumn{1}{c}{Text with specific prefix} &
  Text with {\fontfamily{pcr}\selectfont [HATE]} tokens \\ \hline
\begin{tabular}[c]{@{}l@{}}Hãnh diện về ng thầy có tâm nhất của năm.\\ (\textit{Translated: Proud of the most dedicated}\\ \textit{teacher of the year.})\end{tabular} &
  {\fontfamily{pcr}\selectfont []} &
  \begin{tabular}[c]{@{}l@{}}Hãnh diện về ng thầy có tâm\\ nhất của năm.\end{tabular} &
  {\fontfamily{pcr}\selectfont []} &
  \begin{tabular}[c]{@{}l@{}}hate-spans-detection: Hãnh diện về\\ ng thầy có tâm nhất của năm.\end{tabular} &
  \multicolumn{1}{l}{\begin{tabular}[c]{@{}l@{}}Hãnh diện về ng thầy có tâm nhất \\của năm.\end{tabular}} \\
\begin{tabular}[c]{@{}l@{}}Chương trình \textcolor{red}{ln} gì vậy ? :D\\ (\textit{Translated: What is this \textcolor{red}{pussy} program ? :D})\end{tabular} &
  {\fontfamily{pcr}\selectfont [\textcolor{red}{13,14}]} &
  Chương trình ln gì vậy ? :D :))) &
  {\fontfamily{pcr}\selectfont O O B-T O O O} &
  \begin{tabular}[c]{@{}l@{}}hate-spans-detection: Chương\\ trình ln gì vậy ? :D :)))\end{tabular} &
  \multicolumn{1}{l}{\begin{tabular}[c]{@{}l@{}}Chương trình {\fontfamily{pcr}\selectfont [HATE]}ln{\fontfamily{pcr}\selectfont [HATE]} \\gì vậy ? :D :)))\end{tabular}} \\
\begin{tabular}[c]{@{}l@{}}t \textcolor{blue}{deo} hieu no cuoi \textcolor{orange}{cl me} gi nua\\ (\textit{Translated: I don't \textcolor{blue}{fucking} understand \textcolor{orange}{what}}\\ \textit{\textcolor{orange}{the fuck} he is laughing at})\end{tabular} &
  {\fontfamily{pcr}\selectfont \begin{tabular}[c]{@{}c@{}}[\textcolor{blue}{2,3,4}\\\textcolor{orange}{19,20,21,22,23}]\end{tabular}} &
  t deo hieu no cuoi cl gi nua & {\fontfamily{pcr}\selectfont
  \begin{tabular}[c]{@{}c@{}}O B-T O O O\\ B-T I-T O O\end{tabular} }&
  \begin{tabular}[c]{@{}l@{}}hate-spans-detection: t deo hieu\\ no cuoi cl me gi nua\end{tabular} &
  \multicolumn{1}{l}{\begin{tabular}[c]{@{}l@{}}t {\fontfamily{pcr}\selectfont [HATE]}deo{\fontfamily{pcr}\selectfont [HATE]} hieu no\\ cuoi {\fontfamily{pcr}\selectfont [HATE]}cl me{\fontfamily{pcr}\selectfont [HATE]} \\gi nua\end{tabular}} \\ \hline
\end{tabular}%
}
\caption{Examples from HSD benchmark datasets in experiments featuring diverse input and output data formats aligned with BERT-based and T5-based models.}
\label{tab:data_example}
\end{table}
\end{landscape}